\documentclass[runningheads]{llncs}

\usepackage[colorlinks]{hyperref}
\usepackage{url}
\usepackage{amsmath}
\usepackage{algorithm}
\usepackage{algorithmic}

\usepackage{subfigure}
\usepackage{lipsum}
\usepackage{multirow}
\usepackage{booktabs}
\usepackage{graphicx}
\usepackage[normalem]{ulem}
\usepackage{amsmath}
\usepackage{amssymb}
\usepackage{mathrsfs}
\usepackage{float}
\usepackage[colorinlistoftodos]{todonotes}

\usepackage[misc]{ifsym}

\newcommand{\name}{CLIPMH}
\begin{document}

\title{CLIP Multi-modal Hashing for Multimedia Retrieval}
\titlerunning{\name}
%
\author{Jian Zhu\inst{1,2} \and
Mingkai Sheng\inst{3} \and
Zhangmin Huang\inst{2} \and 
Jingfei Chang\inst{2} \and \\
Jinling Jiang\inst{2} \and
Jian Long\inst{2} \and
Cheng Luo\inst{2} \and
Lei Liu\inst{1,*}}

%
\institute{School of Information Science and Technology, University of Science and Technology of China \\
\email{\{liulei13\}@ustc.edu.cn} \and 
Zhejiang Lab \\
\email{\{qijian.zhu, jiangjinling, cjf\_chang, longjian, zmhuang, luo\_cheng\}@zhejianglab.com} \and
University of Chinese Academy of Sciences \\ 
\email{\{shengmingkai22\}@mails.ucas.ac.cn} 
\\}

\toctitle{CLIPMH}
\tocauthor{}
\authorrunning{Jian Zhu et al.}

%

%
%

%

\maketitle
\def\thefootnote{*}\footnotetext{Corresponding author.}

\begin{abstract}
Multi-modal hashing methods are widely used in multimedia retrieval, which can fuse multi-source data to generate binary hash code. However, the individual backbone networks have limited feature expression capabilities and are not jointly pre-trained on large-scale unsupervised multi-modal data, resulting in low retrieval accuracy. To address this issue, we propose a novel  \textit{CLIP Multi-modal Hashing} (CLIPMH) method. Our method employs the CLIP framework to extract both text and vision features and then fuses them to generate hash code. Due to enhancement on each modal feature, our method has great improvement in the retrieval performance of multi-modal hashing methods. Compared with state-of-the-art unsupervised and supervised multi-modal hashing methods, experiments reveal that the proposed CLIPMH  can significantly improve performance (a maximum increase of $8.38\%$ in mAP).
\end{abstract}
\section{Introduction}

Multi-modal hashing \cite{zhu:24,zhu:25,zhu:26,zhu:27} is one of the important technologies in the field of multimedia retrieval, which fuses multi-modal heterogeneous data to generate hash codes. Since the backbone networks lack good feature expression capability, the current multi-modal hashing methods suffer from the problem of low retrieval accuracy. For instance, Flexible Multi-modal Hashing (FDH) \cite{zhu:52} and Bit-aware Semantic Transformer Hashing (BSTH) \cite{tan:61} hire a VGGNet \cite{simonyan:51} for the vision modal and a Bag-of-Words model \cite{zhang:53} for the text modal. Further, Deep Metric Multi-View Hashing (DMMVH) \cite{zhu:62} uses Deep ResNet \cite{he:5} and BERT-base \cite{devlin:3} as backbone networks. The present backbones have become antiquated in the context of feature extraction, therefore, this necessitates an update in feature extraction methods. Moreover, these backbone networks which operate within individual modalities are not trained jointly, thus, there is no semantic alignment. Consequently, the semantic gap remains unresolved. This degrades the overall retrieval accuracy of current multi-modal hashing techniques.
\begin{figure}
  \centering
  \includegraphics[width=8cm]{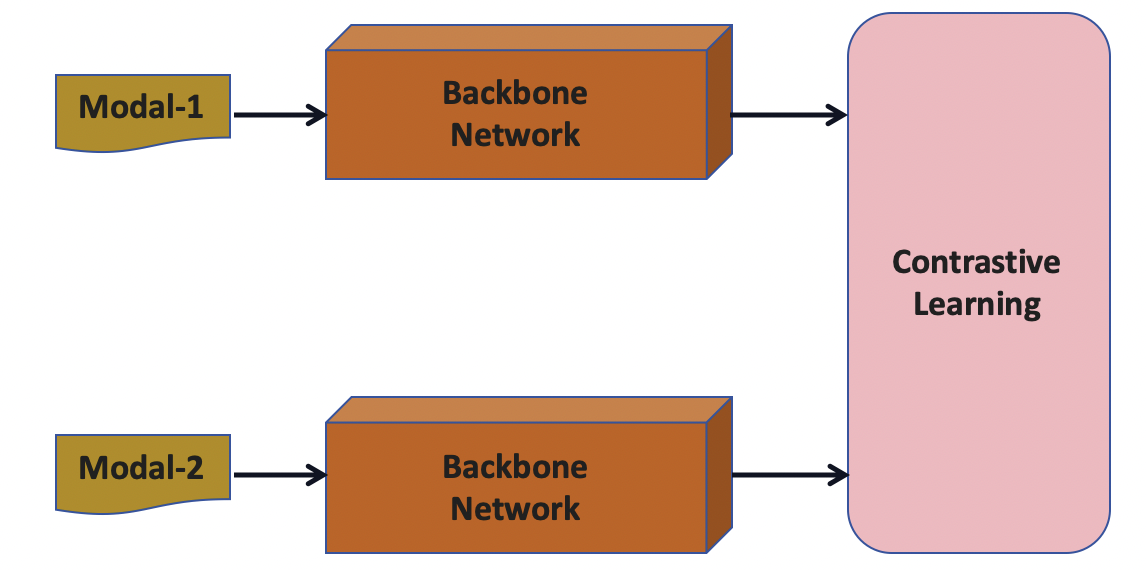}
  \caption{A schematic illustration of CLIP framework. A standard vision backbone network and a text backbone network extract vision (Modal-1) and text (Modal-2) features, respectively. With these two backbones, the CLIP framework learns a vision-text pair through contrastive learning.}
  \label{fig:01}
\end{figure}

In recent years, multi-modal large-scale models have made significant advancements, attributed to their training on extensive data, endowing them with robust semantic expressive capabilities. A notable example of such models is Contrastive Language-Image Pre-training (CLIP) \cite{radford:63}. Surprisingly, the incorporation of these large-scale models into the area of multi-modal retrieval has not been explored extensively. Here, for the first time, we delve into the impact of CLIP on the efficiency of multi-modal hashing for retrieval. As illustrated in Fig. \ref{fig:01}, CLIP undergoes contrastive learning on a large collection of vision-text pairs, exhibiting remarkable ability for zero-shot and few-shot learning, as well as profound semantic comprehension. Its significant impact on the multi-modal domain has been recognized, but a thorough evaluation of its impact and effectiveness in multi-modal hashing retrieval has yet to be carried out.
\begin{figure*}
  \centering
  \includegraphics[width=12cm]{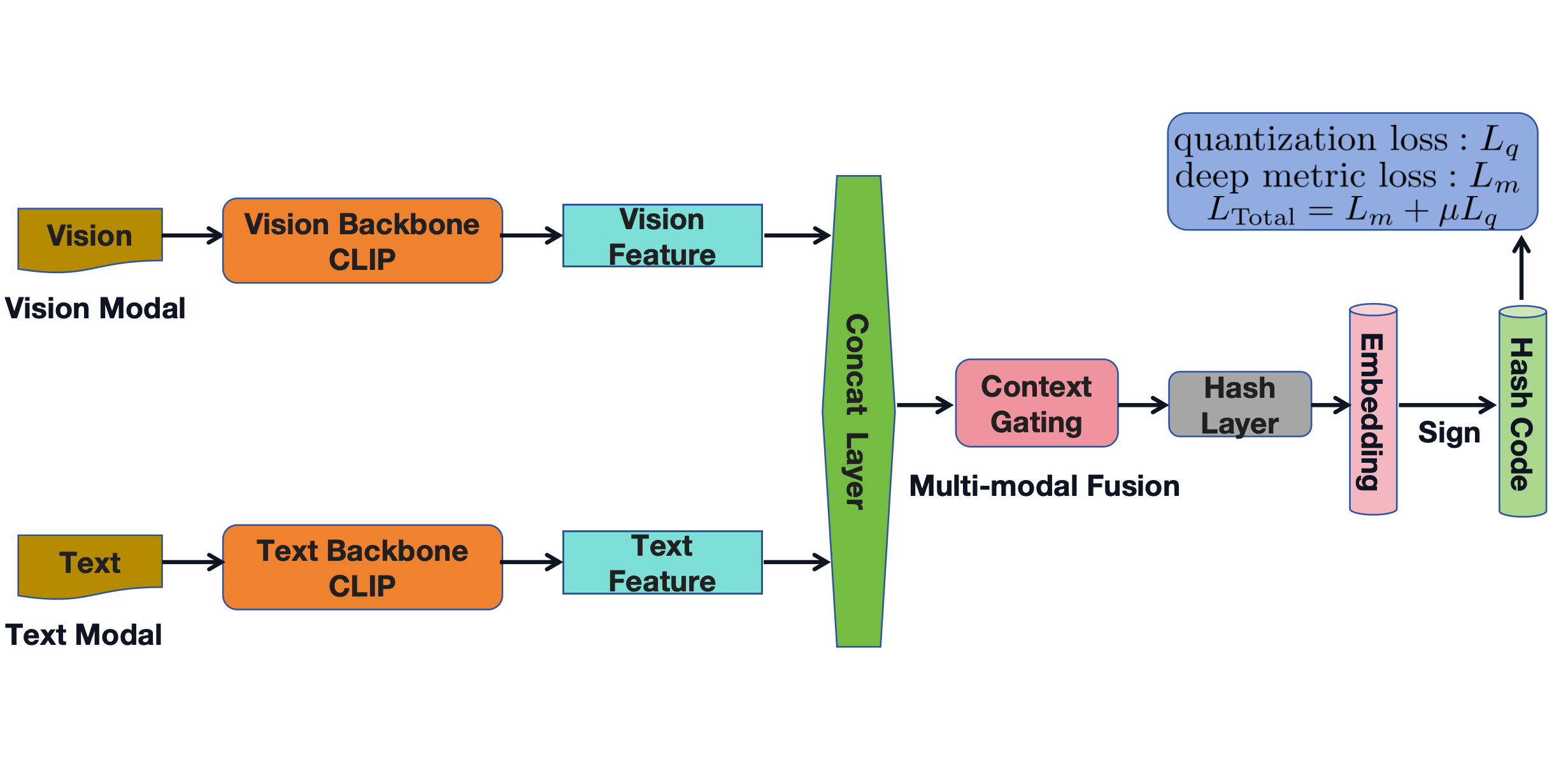}
  \caption{A schematic illustration of the proposed CLIPMH method for Multi-modal retrieval. The Vision-/Text-CLIP backbones extract vision/text features, respectively, and then these two individual features are passed through the concatenation layer before feeding them into the Multi-modal Fusion module. Finally, based on the fused features, the Hash layer generates the hash codes.}
  \label{fig:02} 
\end{figure*}
To remedy these issues, this paper proposes a novel \textit{CLIP Multi-modal Hashing} method termed CLIPMH. Our work explores the untapped potential of CLIP in the realm of multi-modal hashing retrieval. Specifically, we leverage the CLIP framework to extract features from both textual and visual domains, resulting in superior modal features. This, in turn, significantly enhances the retrieval performance of extant multi-modal hashing methods. In contrast to the most recent state-of-the-art method, our proposed CLIPMH demonstrates a remarkable improvement of up to $8.38\%$. 

Our contributions are summarized as follows:
\begin{itemize}
\item To address the issue of inadequate semantic representation in the backbone network of multi-modal hashing techniques, we propose a novel multi-modal hashing method termed CLIPMH. To the best of our knowledge, our method pioneers the application of multi-modal large models into the multi-modal hashing retrieval. 
\item More precisely, we utilize the CLIP framework to obtain improved feature representations, and then fusing these better features can help improve the performance of downstream tasks.
\item We conduct extensive experiments on the MIR-Flickr25K, NUS-WIDE, and MS COCO datasets. The CLIPMH achieves state-of-the-art results in multimedia retrieval tasks. 
\end{itemize}

\section{Related Work}
Multi-view hashing \cite{liu:7,kang:8,song:9,liu:10,shen:11,yang:12,kim:13} fuses multi-view data to generate hash representation. These methods use a graph to construct the relationships among different views for hash code. Multiple Feature Hashing (MFH) \cite{song:9} not only preserves the local structure data of each view but also considers the whole data during the optimization. Multi-view Alignment Hashing (MAH) \cite{liu:10} focuses on the hidden semantic information and obtains the joint distribution of the data. Multi-view Discrete Hashing (MvDH) \cite{shen:11} performs spectral clustering to learn cluster centers. Multi-view Latent Hashing (MVLH) \cite{shen:6} produces shared multi-view hash codes from a unified kernel embedding space.  Compact Kernel Hashing with Multiple Features (MFKH) \cite{liu:7} treats supervised multi-view hash representation learning as a similarity preserving problem. Different from MFKH, Discrete Multi-view Hashing (DMVH) \cite{yang:12} creates a similarity graph based on Locally Linear Embedding (LLE)  \cite{hou:14,saul:15}. 

In recent years, some deep multi-view hashing methods have been proposed. Flexible Discrete Multi-view Hashing (FDMH) \cite{liu:23}, Flexible Online Multi-modal Hashing (FOMH) \cite{lu:24}, and Supervised Adaptive Partial Multi-view Hashing (SAPMH) \cite{zheng:25} aims to get a projection from input space to embedding space by using nonlinear methods. The learned embeddings are fused into multi-modal embedding for multi-view hashing methods. Instead of seeking an embedding space, Deep Collaborative Multi-view Hashing (DCMVH) \cite{zhu:17} directly generates hash codes using a deep network. A discriminative dual-level semantic method is proposed for their supervised training. FGCMH \cite{lu:18} is based on a graph convolutional network (GCN). It preserves both the modality-individual and modality-fused structural similarity to produce hash code. To facilitate multi-view hash at the concept aspect, Bit-aware Semantic Transformer Hashing (BSTH) \cite{tan:61} focuses on bit-wise semantic concepts while aligning disparate views. Deep Metric Multi-View Hashing (DMMVH) \cite{zhu:62} effectively fuses the multi-view features and utilizes the metric information provided by the dissimilar samples.

\section{The Proposed Methodology}
To address the limited expressive power of backbone networks, we propose CLIPMH which incorporates the CLIP framework. The encoder backbone of CLIP exhibits robust proficiency. This confers significant benefits to multi-modal hashing.
\subsection{Deep Multi-modal Hashing Network}
\label{section:proposed_method} 
We propose Deep Multi-modal Hashing Network to generate hash codes, which mainly consists of three parts: (1) CLIP Backbone, (2) Multi-modal Fusion Module, and (3) Hash Layer. Each part will be detailed separately in the following.

\textbf{CLIP Backbone.} The \textit{encode-vision} module of CLIP \cite{radford:63} is utilized as the vision backbone for extracting vision features, which is based on ResNet \cite{he:5} or ViT \cite{dosovitskiy:64}. Similar to the vision backbone, the \textit{encode-text} module of CLIP that based on Transformer\cite{vaswani:4} is harnessed to extract corresponding text features. During the training process of CLIP, contrastive learning is used to optimize these two feature extraction modules. Specifically, for each training batch, $N$ vision features and $N$ text features are paired to form an $N\times N$ vision-text matrix. By simultaneously maximizing the similarities of positive samples on the diagonal of the matrix and minimizing the similarities of negative samples off the diagonal, the contrastive learning module can predict the cosine similarity of $N\times N$ vision-text pairs. Through contrastive learning optimization, vision-text alignment is achieved. Meanwhile, benefiting from the contrastive learning and the extensive dataset of vision-text pairs used for training, the \textit{encode-vision} and \textit{encode-text} modules can obtain remarkable capacities for feature representation.
\begin{table*}
  \centering
        \caption{General statistics of three datasets. The dataset size, number of categories, and feature dimensions are included.}
  \resizebox{\textwidth}{!}{\begin{tabular}{llllllll}
    \toprule[1pt]
    Dataset   & Training Size & Retrieval Size & Query Size & Categories&Visual Embedding & Textual Embedding \\ \midrule[0.8pt]
    MIR-Flickr25K & 5000  & 17772 & 2243    & 24&512-D & 512-D \\
  NUS-WIDE & 21000  & 193749 & 2085    & 21&512-D &512-D\\
    MS COCO & 18000  & 82783 & 5981    & 80&512-D &512-D\\
    \bottomrule[1pt]
  \end{tabular}}
  \label{Tab:01}
\end{table*}

\textbf{Multi-modal Fusion Module.}
We employ the Context-Gating \cite{miech:54} method to fuse the concatenated vision and text features. The multi-modal fusion module is represented as:
\begin{equation}
Z_{\text{f}} =\sigma(w_{\text{f}}Z_{\text{c}}+b_{\text{f}})\circ Z_{\text{c}}
\end{equation}
where $Z_{\text{c}} \in \mathbb{R}^{1024}$ represents the concatenated feature vector, $\sigma$ is the non-linear sigmoid activation function, and $\circ$  denotes an element-wise multiplication operation. Given the learnable parameters $w_{\text{f}} \in \mathbb{R}^{1024 \times 1024}$ and $b_{\text{f}} \in \mathbb{R}^{1024}$, a weight vector $\sigma(w_{\text{f}}X_{\text{c}}+b_{\text{f}}) \in [0, 1]$ is obtained, representing the learned gate. Then, the fused feature $Z_{\text{f}}$ is attained by applying this gate to each dimension of the concatenated feature $Z_{\text{c}}$.
 
\textbf{Hash Layer.} The Hash layer transforms the fused features into hash codes, which can be represented as:
\begin{equation}
    h_{\text{k-bit}} = \tanh(w_{\text{h}}Z_{\text{f}}+b_{\text{h}}),
\end{equation}
where $k$ is $k$-bit hash code, $w_{\text{h}} \in \mathbb{R}^{1024 \times k}$  and $b_{\text{h}} \in \mathbb{R}^{k}$ are the learnable parameters.

\subsection{Loss Funtion}
Assume that the training dataset $\mathcal{X}= \left\{\left\{(x_{i}, y_i) \right\}_{i=1}^{N}\right\}$, where $x_{i} \in \mathbb{R}^{D}$ is a multi-modal instance and $y_{i}$ denotes the category information of $x_{i}$. During the training process, the loss function $L_m$ penalizes dissimilar samples that exhibit a closer distance in the embedding space, while simultaneously rewarding larger distances between them, which can be represented as:
\begin{equation}
  L_{m}=\frac{1}{(\lambda b)^2}\sum\limits_{i=1}^{\lambda b}\sum\limits_{j=(1-\lambda)b+1}^{b}[\delta \log (1+e^{{\Theta}_{ij} })-\Phi_{ij} \Theta_{ij}],
  \label{eq:Lm2}
\end{equation} 
where $\Theta_{ij}$ is the inner product of hash code $x_{i}$ and $x_j$, and $\Phi_{ij}$ is defined as follows: if $x_i$ and $x_j$ are semantically similar then $\Phi_{ij} = 1$, otherwise, $\Phi_{ij} = 0$. $\delta$ is a hyper-parameter, which represents the loss weight of dissimilar sample pairs. To reduce the computational complexity, a hyper-parameter $\lambda$ is introduced, and $b$ is the batch size.

Furthermore, the quantization loss $L_q$ is employed to refine the generated hash codes, which is formulated as:
\begin{equation}
L_{q}=\frac{1}{b}\sum_{i\in I} ({\left\|\left|\boldsymbol{h}_{i}\right|-\mathbf{1}\right\|_{2}}),
\end{equation}
where $h_i$ is the hash code of $x_i$, $I= \{i \mid 1\le i \leq \lambda b, i \in \mathbb{N}\}\cup \{i \mid(1-\lambda) b+1\le i \leq b, i \in \mathbb{N}\}$. Combining the metric loss and the quantization loss by weighted sum \cite{zhu:62} yields the total loss function of our method:
\begin{equation}
  L_{\text{Total}}= L_{m}+\mu L_{q},
  \label{eq:loss}
\end{equation}
where $\mu$ is a hyper-parameter obtained through grid search.

To recap, due to the high efficiency of the backbone and well-designed loss functions, our method can extract high-quality multi-modal features and fuse them to an optimal global representation, ultimately generating hash code for multimedia retrieval tasks through the Hash layer.

\section{Experiments}
\subsection{Evaluation Datasets and Metrics}
In the experiments, we evaluate the performance of the proposed CLIPMH on large-scale multimedia retrieval tasks. We utilize three well-known datasets: MIR-Flickr25K \cite{huiskes:20}, NUS-WIDE \cite{chua:22}, and MS COCO \cite{lin:21}. These datasets have gained widespread usage for evaluating the performance of multimedia retrieval systems. The mean Average Precision (mAP) is employed as the evaluation metric. Table \ref{Tab:01} provides a summary of the dataset statistics used in the experiments.

\subsection{Baseline}
To evaluate the retrieval metric, we compare the proposed CLIPMH method with thirteen multi-modal hashing methods, including four unsupervised methods (e.g., Multiple Feature Hashing (MFH) \cite{song:9}, Multi-view Alignment Hashing (MAH) \cite{liu:10}, Multi-view Latent Hashing (MVLH) \cite{shen:6}, and Multi-view Discrete Hashing (MvDH) \cite{shen:11}) and nine supervised methods (e.g., Multiple Feature Kernel Hashing (MFKH) \cite{liu:7}, Discrete Multi-view Hashing (DMVH) \cite{yang:12}, Flexible Discrete Multi-view Hashing (FDMH) \cite{liu:23}, Flexible Online Multi-modal Hashing (FOMH) \cite{lu:24}, Deep Collaborative Multi-View Hashing (DCMVH) \cite{zhu:17}, Supervised Adaptive Partial Multi-view Hashing (SAPMH) \cite{zheng:25}, Flexible Graph Convolutional Multi-modal Hashing (FGCMH) \cite{lu:18}, Bit-aware Semantic Transformer Hashing (BSTH) \cite{tan:61} and Deep Metric Multi-View Hashing (DMMVH) \cite{zhu:62}).

\begin{table*}
  \setlength{\tabcolsep}{2pt}
  \centering
  \caption{The comparable mAP results on MIR-Flickr25K, NUS-WIDE, and MS COCO. The best results are bolded, and the second-best results are underlined. The * indicates that the results of our method on this dataset are statistical significance.}
  \resizebox{\textwidth}{!}{\begin{tabular}{llllllllllllll}
    \toprule[1pt]
    \multicolumn{1}{c}{\multirow{2}{*}{Method}} & \multicolumn{1}{c}{\multirow{2}{*}{Ref.}} & \multicolumn{4}{c}{MIR-Flickr25K*}    & \multicolumn{4}{c}{NUS-WIDE*}      & \multicolumn{4}{c}{MS   COCO*}       \\  \cmidrule(r){3-6}  \cmidrule(r){7-10}  \cmidrule(r){11-14}
    \multicolumn{1}{c}{}                         & \multicolumn{1}{c}{}                      & 16 bits & 32 bits & 64 bits & 128 bits & 16 bits & 32 bits & 64 bits & 128 bits & 16 bits & 32 bits & 64 bits & 128 bits \\ \midrule[0.8pt]
    MFH                                          & TMM13                                     & 0.5795 & 0.5824 & 0.5831 & 0.5836  & 0.3603 & 0.3611 & 0.3625 & 0.3629  & 0.3948 & 0.3699 & 0.3960  & 0.3980   \\
    MAH                                          & TIP15                                     & 0.6488 & 0.6649 & 0.6990  & 0.7114  & 0.4633 & 0.4945 & 0.5381 & 0.5476  & 0.3967 & 0.3943 & 0.3966 & 0.3988  \\
    MVLH                                         & MM15                                      & 0.6541 & 0.6421 & 0.6044 & 0.5982  & 0.4182 & 0.4092 & 0.3789 & 0.3897  & 0.3993 & 0.4012 & 0.4065 & 0.4099  \\
    MvDH                                         & TIST18                                    & 0.6828 & 0.7210  & 0.7344 & 0.7527  & 0.4947 & 0.5661 & 0.5789 & 0.6122  & 0.3978 & 0.3966 & 0.3977 & 0.3998  \\ \midrule[0.8pt]
    MFKH                                         & MM12                                      & 0.6369 & 0.6128 & 0.5985 & 0.5807  & 0.4768 & 0.4359 & 0.4342 & 0.3956  & 0.4216 & 0.4211 & 0.4230  & 0.4229  \\
    DMVH                                         & ICMR17                                    & 0.7231 & 0.7326 & 0.7495 & 0.7641  & 0.5676 & 0.5883 & 0.6902 & 0.6279  & 0.4123 & 0.4288 & 0.4355 & 0.4563  \\
    FOMH                                         & MM19                                      & 0.7557 & 0.7632 & 0.7564 & 0.7705  & 0.6329 & 0.6456 & 0.6678 & 0.6791  & 0.5008 & 0.5148 & 0.5172 & 0.5294  \\
    FDMH                                         & NPL20                                     & 0.7802 & 0.7963 & 0.8094 & 0.8181  & 0.6575 & 0.6665 & 0.6712 & 0.6823  & 0.5404 & 0.5485 & 0.5600   & 0.5674  \\
    DCMVH                                        & TIP20                                     & 0.8097 & 0.8279 & 0.8354 & 0.8467  & 0.6509 & 0.6625 & 0.6905 & 0.7023  & 0.5387 & 0.5427 & 0.5490  & 0.5576  \\
    SAPMH                                        & TMM21                                      & 0.7657 & 0.8098 & 0.8188 & 0.8191  & 0.6503 & 0.6703 & 0.6898 & 0.6901  & 0.5467 & 0.5502 & 0.5563 & 0.5672  \\
    FGCMH                                        & MM21                                      & 0.8173 & 0.8358 & 0.8377 & 0.8606  & 0.6677 & 0.6874 & 0.6936 & 0.7011  & 0.5641 & 0.5273 & 0.5797 & 0.5862 \\
  BSTH & SIGIR22  &0.8145&0.8340&0.8482&0.8571                                     & 0.6990 & 0.7340 & 0.7505 & 0.7704  & 0.5831 & 0.6245 & 0.6459 & 0.6654  \\
  DMMVH                                         & ICME23                                         &\underline{0.8587} & \underline{0.8707} & \underline{0.8798} & \underline{0.8827} & \underline{0.7714} & \underline{0.7820} & \underline{0.7879} & \underline{0.7916}  &\underline{0.6716} & \underline{0.7030} & \underline{0.7122} & \underline{0.7244}\\\midrule[0.8pt]
    CLIPMH                                         & ours                                         &\textbf{0.8862} & \textbf{0.8921} & \textbf{0.8997} & \textbf{0.9018} & \textbf{0.7802} & \textbf{0.7986} & \textbf{0.8029} & \textbf{0.8085}  &\textbf{0.6806} & \textbf{0.7450} & \textbf{0.7693} & \textbf{0.8082} \\
    \bottomrule[1pt]
  \end{tabular}}

  \label{Tab:02}
\end{table*}

\subsection{Analysis of Experimental Results}

\begin{table*}[!t]
  \setlength{\tabcolsep}{2pt}
  \centering
        \caption{Ablation Experiments On Three Datasets. Effects of Deep Multi-modal Hash Network Architecture.}
  \resizebox{\textwidth}{!}{\begin{tabular}{lllllllllllll}
    \toprule[1pt]
    \multicolumn{1}{c}{\multirow{2}{*}{Methods}} & \multicolumn{4}{c}{MIR-Flickr25K}   & \multicolumn{4}{c}{NUS-WIDE}  & \multicolumn{4}{c}{MS COCO} \\   \cmidrule(r){2-5}  \cmidrule(r){6-9}  \cmidrule(r){10-13}
    \multicolumn{1}{c}{} & 16 bits & 32 bits & 64 bits & 128 bits & 16 bits & 32 bits & 64 bits & 128 bits& 16 bits & 32 bits & 64 bits & 128 bits \\  \midrule[0.8pt] 
    CLIPMH-text    & 0.7124 &  0.7316 &  0.7371 &  0.7384  & 0.6789 & 0.6920 & 0.6968 & 0.6994 &0.6527&0.7154&0.7327&0.7540\\
    CLIPMH-image    & 0.8552 &  0.8745 &  0.8840 &  0.8923  & 0.7681 & 0.7711 &0.7831 & 0.7897 & 0.6589&0.7205&0.7417&0.7658    \\
    CLIPMH-concat    & 0.8634 &  0.8820 &  0.8896 &  0.8966  & 0.7756 & 0.7877 & 0.7943 & 0.8011   & 0.6708&0.7384&0.7610&0.7899 \\ 
    ResNet\&BERT &0.8590 & 0.8712 & 0.8792 & 0.8819 & 0.7726 & 0.7818 & 0.7882 & 0.7914  &0.6721 & 0.7028 & 0.7130 & 0.7238 \\\midrule[0.8pt]
    CLIPMH     &\textbf{0.8862} & \textbf{0.8921} & \textbf{0.8997} & \textbf{0.9018} & \textbf{0.7802} & \textbf{0.7986} & \textbf{0.8029} & \textbf{0.8085}  &\textbf{0.6806} & \textbf{0.7450} & \textbf{0.7693} & \textbf{0.8082} \\
    \bottomrule[1pt]
  \end{tabular}}
  
  \label{Tab:03}
\end{table*}
The experimental results are presented in Table \ref{Tab:02}. We can observe that the proposed CLIPMH outperforms all the compared multi-modal hashing methods by a significant margin. Specifically, when compared to the current state-of-the-art multi-modal hashing method  DMMVH \cite{zhu:62}, our approach achieves average mAP improvements of $2.51\%$, $1.83\%$, and $6.82\%$ on the MIR-Flickr25K, NUS-WIDE, and MS COCO datasets, respectively. These superior results are attributed to the introduction of the CLIP framework, which extracts vision/text features and enhances semantic representation.

\subsection{Ablation Studies}
We carry out ablation studies with various experiment conditions, then we present the results to assess the CLIPMH method. The experiment settings are described as follows:
\begin{itemize}
  \item \emph{CLIPMH-vision}: Retrieval relies only on the extracted vision features.
  \item \emph{CLIPMH-text}: For retrieval, just the extracted text embedding is utilized.
  \item \emph{CLIPMH-concat}: Vision and text features are fused with concatenation.
 \item \emph{ResNet\&BERT}: Deep ResNet \cite{he:5} is employed to produce vision features. The BERT-base \cite{devlin:3} is utilized to extract text features.
  \item \emph{CLIPMH}: Our full method.
\end{itemize}
All ablation experiments use the same loss functions. The performance of CLIPMH-vision is better than that of CLIPMH-text, indicating that the vision features are better than the text features. Comparing the two experimental settings of CLIPMH-text and CLIPMH-vision, CLIPHM-concat has improved upon both. Moreover, compared with previous multi-modal hashing methods, CLIPMH-concat achieves state-of-the-art results. Obviously, a simple combination of vision and text features is greatly more effective than baseline methods. This also demonstrates the superiority of our method.

The mAP of CLIPMH outperforms that of ResNet\&BERT, indicating that features extracted by CLIP are better than those of the combination of ResNet and BERT. The comparison between CLIPMH and CLIPMH-concat depicts that the multi-modal fusion module further improves the performance of hash retrieval. Through the above ablation analysis, we can conclude that the CLIP framework is very successful in multi-modal hash tasks.

\begin{figure}
  \centering
  \subfigure{\includegraphics[scale=0.3]{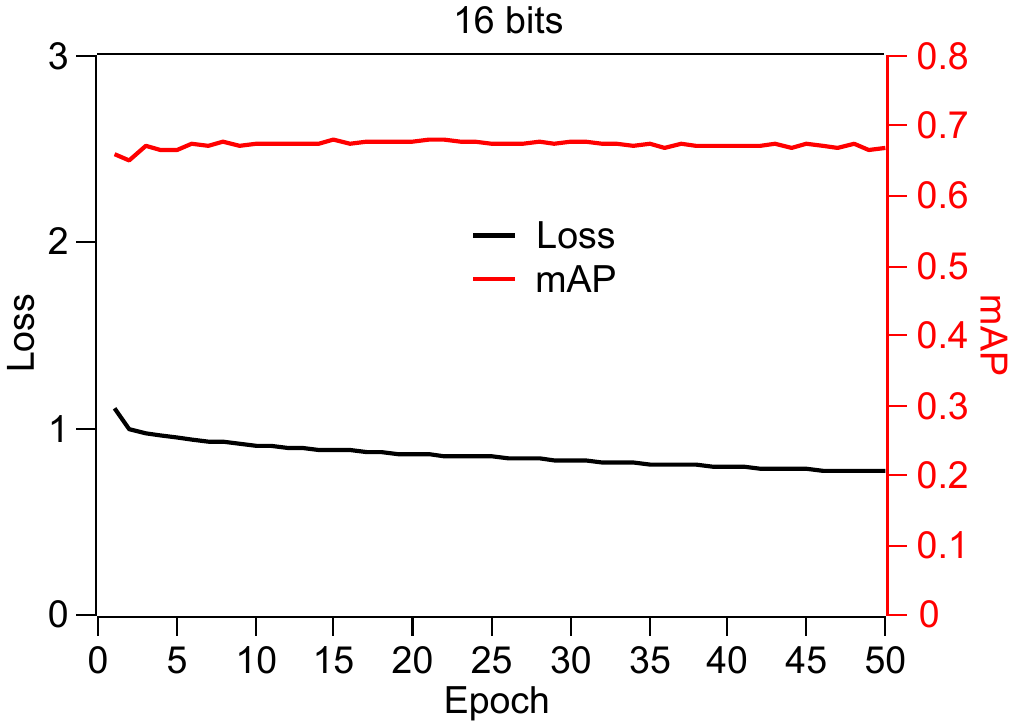}}
  \subfigure{\includegraphics[scale=0.3]{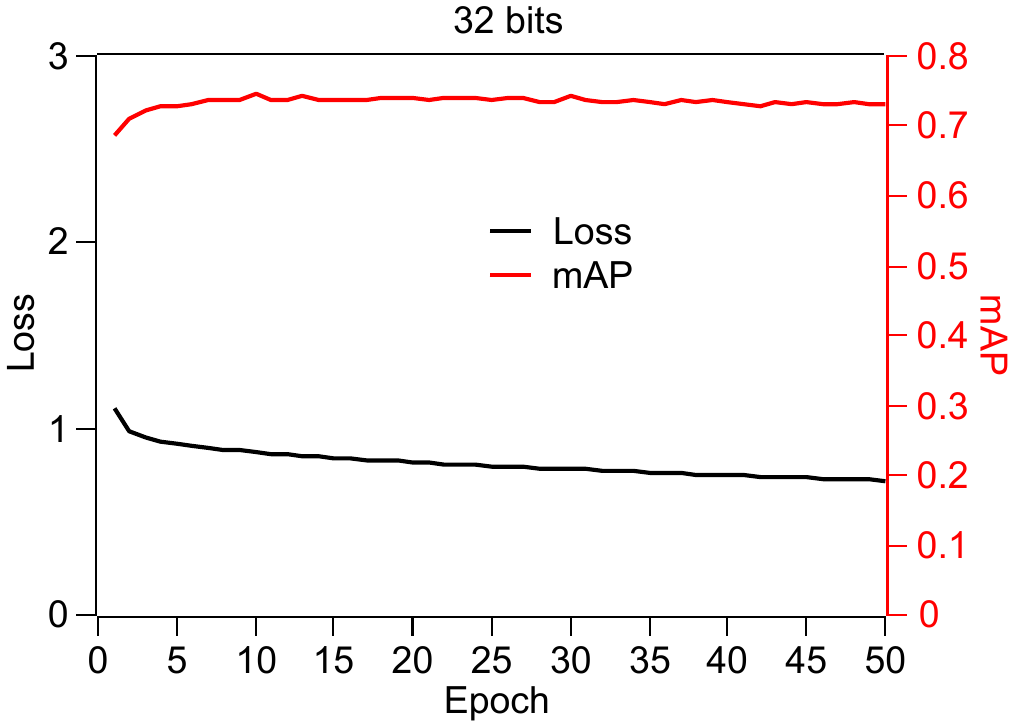}}
  \subfigure{\includegraphics[scale=0.3]{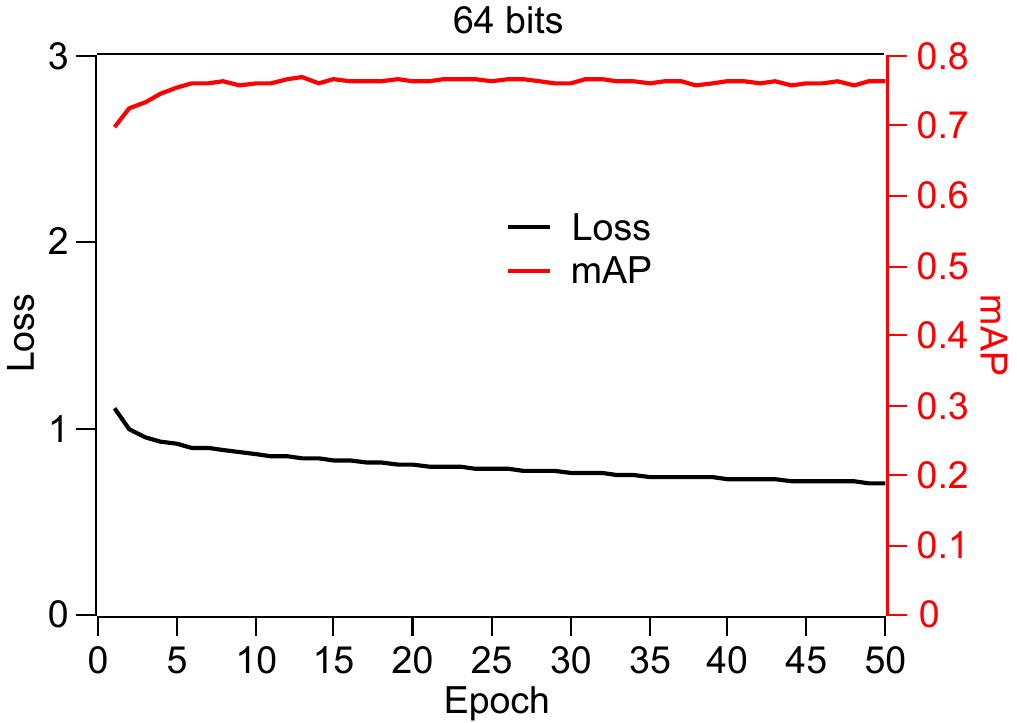}}
  \subfigure{\includegraphics[scale=0.3]{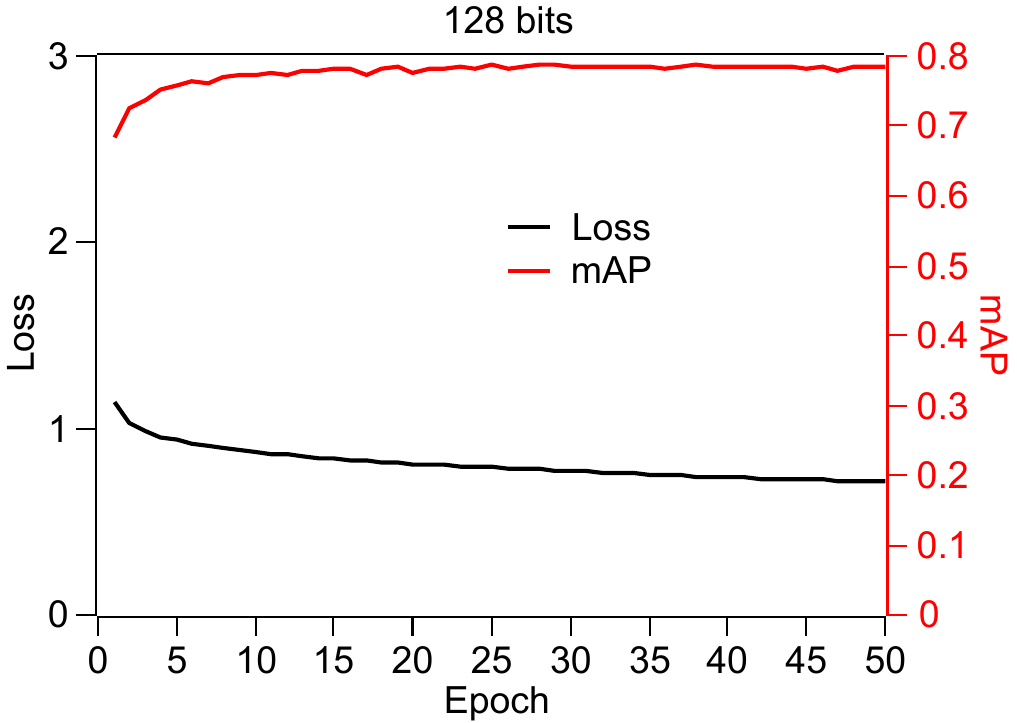}}
  \caption{The training loss and test mAP curves on MS COCO dataset.}
  \label{fig:03}
\end{figure}

\begin{figure}
  \centering
  \subfigure{\includegraphics[scale=0.3]{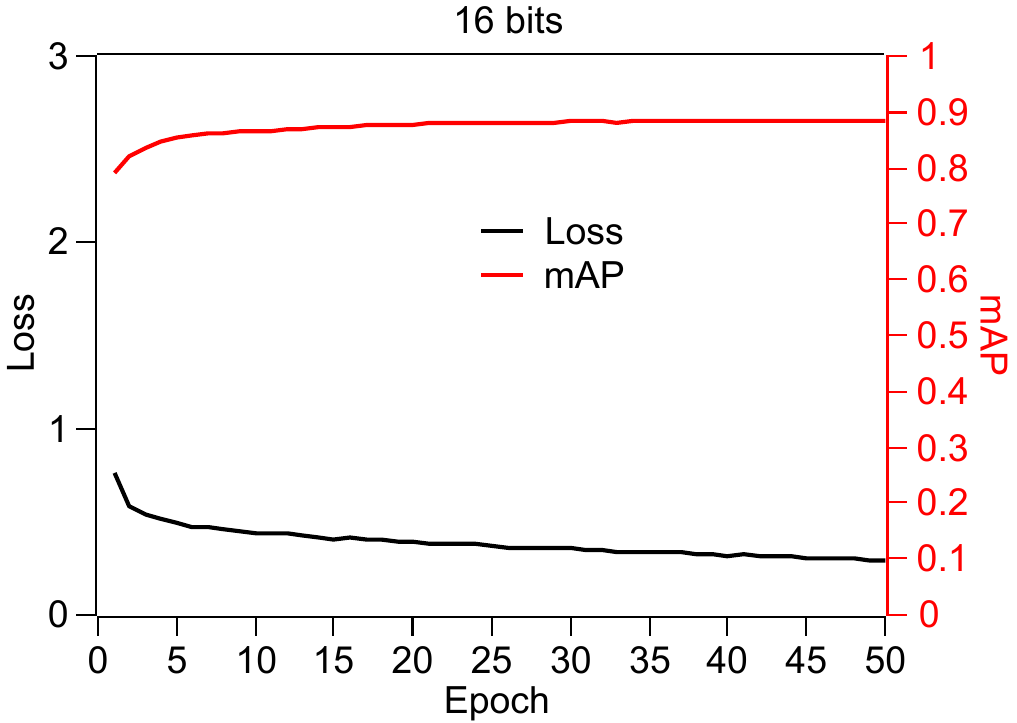}}
  \subfigure{\includegraphics[scale=0.3]{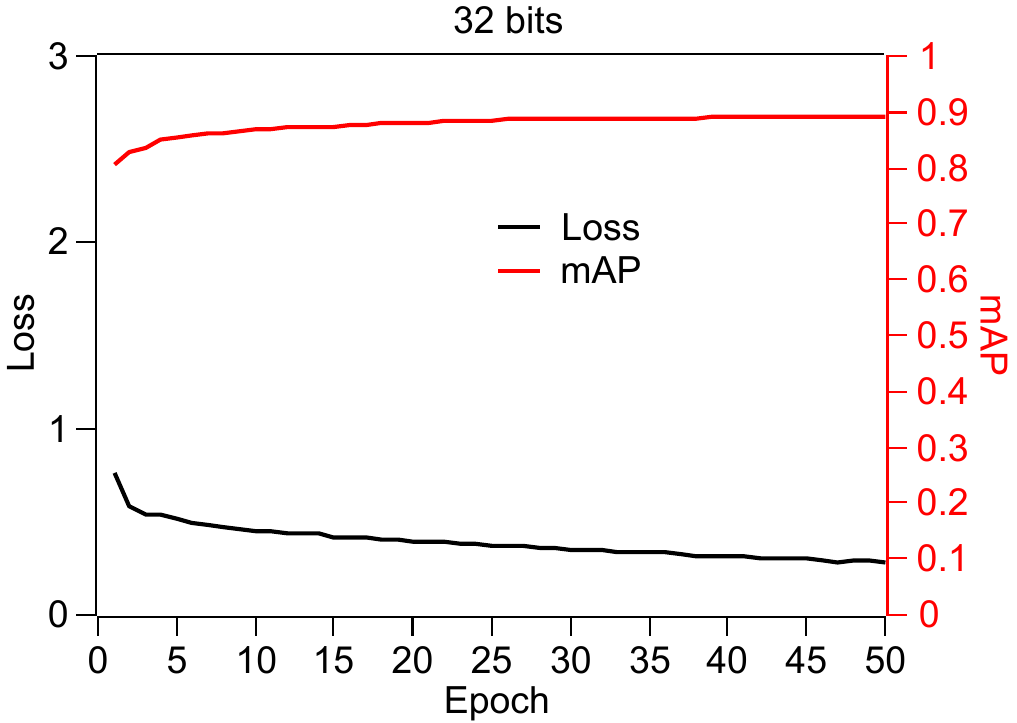}}
  \subfigure{\includegraphics[scale=0.3]{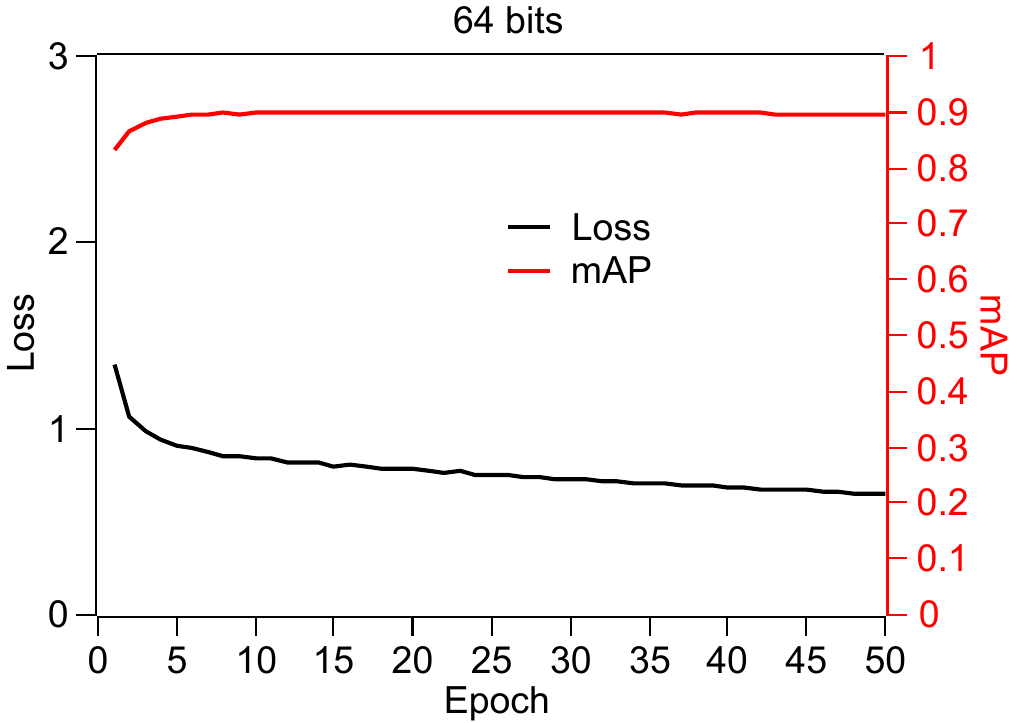}}
  \subfigure{\includegraphics[scale=0.3]{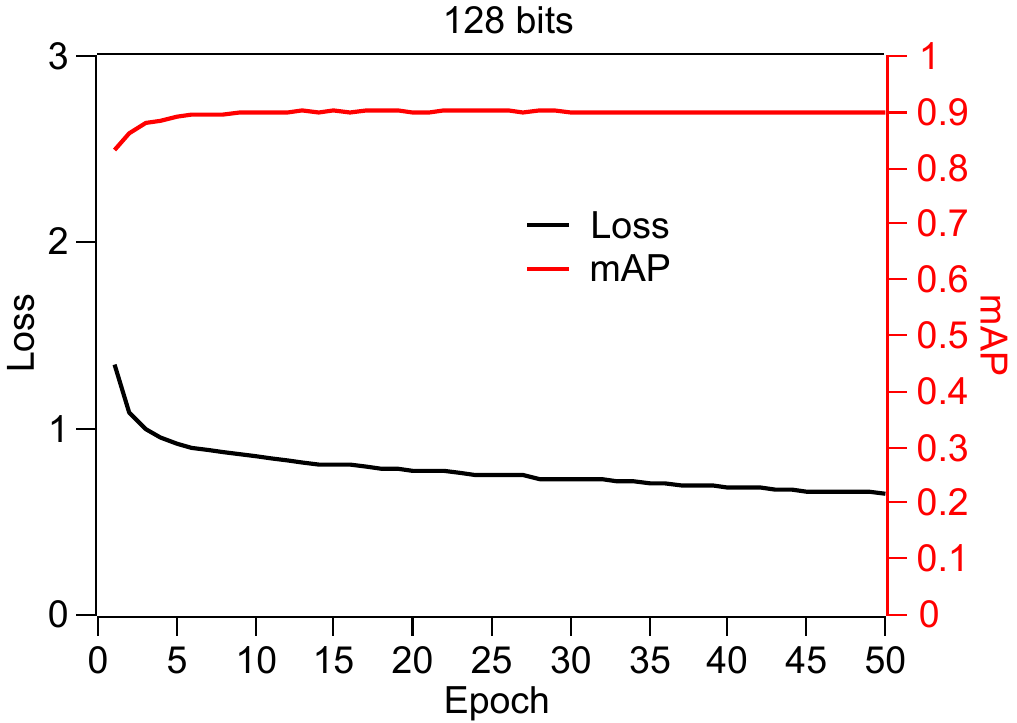}}
  \caption{The training loss and test mAP curves on MIR-Flickr25K dataset.}
  \label{fig:04}
\end{figure}

\subsection{Convergence Analysis}
To verify the convergence properties of CLIPMH, more experiments are carried out. We evaluate hash benchmarks with various code lengths on the MS COCO dataset. Fig. \ref{fig:03}  and Fig. \ref{fig:04} show the training loss and test mAP curves on different datasets. It can be observed that the loss steadily decreases during training. The loss stabilizes after 45 epochs, demonstrating that the local minimum is reached. In contrast, the test mAP rapidly increases at the beginning of training. After 10 epochs, it remains stable. Further training does not result in deterioration of the test mAP, suggesting excellent generalization ability without overfitting. Similar convergence results can be observed for other datasets as well. 

\section{Conclusion and Future Work}
To improve the retrieval accuracy, we propose a novel multi-modal hashing method termed CLIPMH. To elaborate, the CLIP framework is applied to extract vision and text features sufficiently in the multi-modal hashing method. Extensive experiments demonstrate that our method achieves state-of-the-art results. In the future, we will explore more applications of multi-modal large models in the field of multimedia retrieval.

\section*{Acknowledgment}
This research is supported by the Anhui Provincial Natural Science Foundation (No. 2408085QF214), the Fundamental Research Funds for the Central Universities (No. WK2100000045), and the National Key R\&D Program of China (2022YFB4500405).



\bibliography{main}

\begin{thebibliography}{10}

\bibitem{zhu:24}
Fast metric multi-view hashing for multimedia retrieval.
\newblock {\em Information Fusion}, 103:102130, 2024.

\bibitem{zhu:25}
Jian Zhu, Yu~Cui, Zhangmin Huang, Xingyu Li, Lei Liu, Lingfang Zeng, and
  Li-Rong Dai.
\newblock Adaptive confidence multi-view hashing for multimedia retrieval.
\newblock In {\em ICASSP 2024 - 2024 IEEE International Conference on
  Acoustics, Speech and Signal Processing (ICASSP)}, pages 7900--7904, 2024.

\bibitem{zhu:26}
Jian Zhu, Wen Cheng, Yu~Cui, Chang Tang, Yuyang Dai, Yong Li, and Lingfang
  Zeng.
\newblock Central similarity multi-view hashing for multimedia retrieval.
\newblock In {\em Web and Big Data}, pages 486--500. Springer Nature Singapore,
  2024.

\bibitem{zhu:27}
Jian Zhu, Zhangmin Huang, Lei Liu, Chang Tang, and Li-Rong Dai.
\newblock Boosted curriculum multi-view hashing for multimedia retrieval.
\newblock {\em IEEE Signal Processing Letters}, 31:2065--2069, 2024.

\bibitem{zhu:52}
Lei Zhu, Xu~Lu, Zhiyong Cheng, Jingjing Li, and Huaxiang Zhang.
\newblock Flexible multi-modal hashing for scalable multimedia retrieval.
\newblock {\em ACM Transactions on Intelligent Systems and Technology (TIST)},
  11(2):1--20, 2020.

\bibitem{tan:61}
Wentao Tan, Lei Zhu, Weili Guan, Jingjing Li, and Zhiyong Cheng.
\newblock Bit-aware semantic transformer hashing for multi-modal retrieval.
\newblock In {\em Proceedings of the 45th International ACM SIGIR Conference on
  Research and Development in Information Retrieval}, pages 982--991, 2022.

\bibitem{simonyan:51}
Karen Simonyan and Andrew Zisserman.
\newblock Very deep convolutional networks for large-scale image recognition.
\newblock {\em arXiv preprint arXiv:1409.1556}, 2014.

\bibitem{zhang:53}
Yin Zhang, Rong Jin, and Zhi-Hua Zhou.
\newblock Understanding bag-of-words model: a statistical framework.
\newblock {\em International journal of machine learning and cybernetics},
  1(1):43--52, 2010.

\bibitem{zhu:62}
Jian Zhu, Zhangmin Huang, Xiaohu Ruan, Yu~Cui, Yongli Cheng, and Lingfang Zeng.
\newblock Deep metric multi-view hashing for multimedia retrieval.
\newblock {\em arXiv preprint arXiv:2304.06358}, 2023.

\bibitem{he:5}
Kaiming He, Xiangyu Zhang, Shaoqing Ren, and Jian Sun.
\newblock Deep residual learning for image recognition.
\newblock In {\em Proceedings of the IEEE conference on computer vision and
  pattern recognition}, pages 770--778, 2016.

\bibitem{devlin:3}
Jacob Devlin, Ming-Wei Chang, Kenton Lee, and Kristina Toutanova.
\newblock Bert: Pre-training of deep bidirectional transformers for language
  understanding.
\newblock {\em arXiv preprint arXiv:1810.04805}, 2018.

\bibitem{radford:63}
Alec Radford, Jong~Wook Kim, Chris Hallacy, Aditya Ramesh, Gabriel Goh,
  Sandhini Agarwal, Girish Sastry, Amanda Askell, Pamela Mishkin, Jack Clark,
  et~al.
\newblock Learning transferable visual models from natural language
  supervision.
\newblock In {\em International conference on machine learning}, pages
  8748--8763. PMLR, 2021.

\bibitem{liu:7}
Xianglong Liu, Junfeng He, Di~Liu, and Bo~Lang.
\newblock Compact kernel hashing with multiple features.
\newblock In {\em Proceedings of the 20th ACM international conference on
  multimedia}, pages 881--884, 2012.

\bibitem{kang:8}
Yoonseop Kang, Saehoon Kim, and Seungjin Choi.
\newblock Deep learning to hash with multiple representations.
\newblock In {\em 2012 IEEE 12th International Conference on Data Mining},
  pages 930--935. IEEE, 2012.

\bibitem{song:9}
Jingkuan Song, Yi~Yang, Zi~Huang, Heng~Tao Shen, and Jiebo Luo.
\newblock Effective multiple feature hashing for large-scale near-duplicate
  video retrieval.
\newblock {\em IEEE Transactions on Multimedia}, 15(8):1997--2008, 2013.

\bibitem{liu:10}
Li~Liu, Mengyang Yu, and Ling Shao.
\newblock Multiview alignment hashing for efficient image search.
\newblock {\em IEEE Transactions on image processing}, 24(3):956--966, 2015.

\bibitem{shen:11}
Xiaobo Shen, Fumin Shen, Li~Liu, Yun-Hao Yuan, Weiwei Liu, and Quan-Sen Sun.
\newblock Multiview discrete hashing for scalable multimedia search.
\newblock {\em ACM Transactions on Intelligent Systems and Technology (TIST)},
  9(5):1--21, 2018.

\bibitem{yang:12}
Rui Yang, Yuliang Shi, and Xin-Shun Xu.
\newblock Discrete multi-view hashing for effective image retrieval.
\newblock In {\em Proceedings of the 2017 ACM on international conference on
  multimedia retrieval}, pages 175--183, 2017.

\bibitem{kim:13}
Saehoon Kim and Seungjin Choi.
\newblock Multi-view anchor graph hashing.
\newblock In {\em 2013 IEEE International Conference on Acoustics, Speech and
  Signal Processing}, pages 3123--3127. IEEE, 2013.

\bibitem{shen:6}
Xiaobo Shen, Fumin Shen, Quan-Sen Sun, and Yun-Hao Yuan.
\newblock Multi-view latent hashing for efficient multimedia search.
\newblock In {\em Proceedings of the 23rd ACM international conference on
  Multimedia}, pages 831--834, 2015.

\bibitem{hou:14}
Yuexian Hou, Peng Zhang, Xingxing Xu, Xiaowei Zhang, and Wenjie Li.
\newblock Nonlinear dimensionality reduction by locally linear inlaying.
\newblock {\em IEEE transactions on neural networks}, 20(2):300--315, 2009.

\bibitem{saul:15}
Lawrence~K Saul and Sam~T Roweis.
\newblock Think globally, fit locally: unsupervised learning of low dimensional
  manifolds.
\newblock {\em Journal of machine learning research}, 4(Jun):119--155, 2003.

\bibitem{liu:23}
Luyao Liu, Zheng Zhang, and Zi~Huang.
\newblock Flexible discrete multi-view hashing with collective latent feature
  learning.
\newblock {\em Neural Processing Letters}, 52(3):1765--1791, 2020.

\bibitem{lu:24}
Xu~Lu, Lei Zhu, Zhiyong Cheng, Jingjing Li, Xiushan Nie, and Huaxiang Zhang.
\newblock Flexible online multi-modal hashing for large-scale multimedia
  retrieval.
\newblock In {\em Proceedings of the 27th ACM international conference on
  multimedia}, pages 1129--1137, 2019.

\bibitem{zheng:25}
Chaoqun Zheng, Lei Zhu, Zhiyong Cheng, Jingjing Li, and An-An Liu.
\newblock Adaptive partial multi-view hashing for efficient social image
  retrieval.
\newblock {\em IEEE Transactions on Multimedia}, 23:4079--4092, 2020.

\bibitem{zhu:17}
Lei Zhu, Xu~Lu, Zhiyong Cheng, Jingjing Li, and Huaxiang Zhang.
\newblock Deep collaborative multi-view hashing for large-scale image search.
\newblock {\em IEEE Transactions on Image Processing}, 29:4643--4655, 2020.

\bibitem{lu:18}
Xu~Lu, Lei Zhu, Li~Liu, Liqiang Nie, and Huaxiang Zhang.
\newblock Graph convolutional multi-modal hashing for flexible multimedia
  retrieval.
\newblock In {\em Proceedings of the 29th ACM International Conference on
  Multimedia}, pages 1414--1422, 2021.

\bibitem{dosovitskiy:64}
Alexey Dosovitskiy, Lucas Beyer, Alexander Kolesnikov, Dirk Weissenborn,
  Xiaohua Zhai, Thomas Unterthiner, Mostafa Dehghani, Matthias Minderer, Georg
  Heigold, Sylvain Gelly, et~al.
\newblock An image is worth 16x16 words: Transformers for image recognition at
  scale.
\newblock {\em arXiv preprint arXiv:2010.11929}, 2020.

\bibitem{vaswani:4}
Ashish Vaswani, Noam Shazeer, Niki Parmar, Jakob Uszkoreit, Llion Jones,
  Aidan~N Gomez, {\L}ukasz Kaiser, and Illia Polosukhin.
\newblock Attention is all you need.
\newblock {\em Advances in neural information processing systems}, 30, 2017.

\bibitem{miech:54}
Antoine Miech, Ivan Laptev, and Josef Sivic.
\newblock Learnable pooling with context gating for video classification.
\newblock {\em arXiv preprint arXiv:1706.06905}, 2017.

\bibitem{huiskes:20}
Mark~J Huiskes and Michael~S Lew.
\newblock The mir flickr retrieval evaluation.
\newblock In {\em Proceedings of the 1st ACM international conference on
  Multimedia information retrieval}, pages 39--43, 2008.

\bibitem{chua:22}
Tat-Seng Chua, Jinhui Tang, Richang Hong, Haojie Li, Zhiping Luo, and Yantao
  Zheng.
\newblock Nus-wide: a real-world web image database from national university of
  singapore.
\newblock In {\em Proceedings of the ACM international conference on image and
  video retrieval}, pages 1--9, 2009.

\bibitem{lin:21}
Tsung-Yi Lin, Michael Maire, Serge Belongie, James Hays, Pietro Perona, Deva
  Ramanan, Piotr Doll{\'a}r, and C~Lawrence Zitnick.
\newblock Microsoft coco: Common objects in context.
\newblock In {\em European conference on computer vision}, pages 740--755.
  Springer, 2014.

\end{thebibliography}

\end{document}